\documentclass[conference]{IEEEtran}
\IEEEoverridecommandlockouts
\usepackage{cite}
\usepackage{amsmath,amssymb,amsfonts}
\usepackage{algorithmic}
\usepackage{graphicx}
\usepackage{textcomp}
\usepackage{xcolor}
\usepackage{booktabs}
\usepackage{multirow}
\def\BibTeX{{\rm B\kern-.05em{\sc i\kern-.025em b}\kern-.08em T\kern-.1667em\lower.7ex\hbox{E}\kern-.125emX}}
\usepackage{bm}
\usepackage{booktabs}
\usepackage{multirow}

\newcommand{\header}[1]{\subsubsection{#1}}

\newcommand{\std}[1]{\scriptsize{$\pm$}}

\newcommand{\name}{CoGenT}

\begin{document}

\title{
% Self-Supervised Representation Learning for Time Series Classification with Contrastive Autoencoder\\
% Contrastive Masked Autoencoder for Time Series Classification\\
% TS-GC: Generative-Contrastive Representation Learning for Time Series Classification\\
% \name: Fusing Contrastive and Generative Representation Learning for Time Series Classification\\
% \name: 
A Unified Contrastive-Generative Framework for 
% Self-Supervised 
Time Series Classification
}
% https://cis.ieee.org/publications/ieee-transactions-on-artificial-intelligence

% \author{\IEEEauthorblockN{1\textsuperscript{st} Anonymous Authors}}
\author{\IEEEauthorblockN{1\textsuperscript{st} Ziyu Liu}
\IEEEauthorblockA{ 
\textit{RMIT}\\
ziyu.liu2@student.rmit.edu.au
}
\and
\IEEEauthorblockN{2\textsuperscript{th} Azadeh Alavi}
\IEEEauthorblockA{
\textit{RMIT}\\
azadeh.alavi@rmit.edu.au}
\and
\IEEEauthorblockN{3\textsuperscript{th} Minyi Li}
\IEEEauthorblockA{
\textit{RMIT}\\
 minyi.li2@rmit.edu.au}
\and
\IEEEauthorblockN{4\textsuperscript{th} Xiang Zhang}
\IEEEauthorblockA{
\textit{UNC Charlotte}\\
xiang.zhang@charlotte.edu}
}

% \author{
% Ziyu Liu$^1$
% \and
% Azadeh Alavi$^1$\and
% Minyi Li \And
% Xiang Zhang$^2$
% \affiliations
% $^1$RMIT, 
% $^2$University of North Carolina, Charlotte\\
% \emails
% ziyu.liu2@student.rmit.edu.au,
% \{azadeh.alavi, minyi.li2\}@rmit.edu.au, 
% xiang.zhang@charlotte.edu
% }

\maketitle

\begin{abstract}
Self-supervised learning (SSL) for multivariate time series mainly includes two paradigms: contrastive methods that excel at instance discrimination and generative approaches that model data distributions. While effective individually, their complementary potential remains unexplored. We propose a \textbf{Co}ntrastive \textbf{Gen}erative \textbf{T}ime series framework (\name), the first framework to unify these paradigms through joint contrastive-generative optimization. 
\name\ addresses fundamental limitations of both approaches: it overcomes contrastive learning's sensitivity to high intra-class similarity in temporal data while reducing generative methods' dependence on large datasets. 
We evaluate \name\ on six diverse time series datasets. The results show consistent improvements, with up to 59.2\% and 14.27\% F1 gains over standalone SimCLR and MAE, respectively. 
Our analysis reveals that the hybrid objective preserves discriminative power while acquiring generative robustness. 
These findings establish a foundation for hybrid SSL in temporal domains. All the code will be released for reproducibility.  
\end{abstract}


\section{Introduction}
Self-supervised learning (SSL) has emerged as a powerful paradigm for learning representations from unlabeled data, particularly in domains where labeled data is scarce or expensive to obtain~\cite{liu2021self}. Specifically, in time series analysis, SSL has gained significant attention due to its ability to leverage the inherent structure and temporal dependencies in sequential data. 
The prominent SSL approaches in time series contain autoregressive methods, contrastive methods, and generative (reconstructive) methods~\cite{zhang2024self}. The autoregressive method leverages the correlation among time series values and their history; the input and output are in the same semantic space, which is more suitable for forecasting but not necessarily good for classification. In time series forecasting, the input and output have the same semantic meanings (e.g., using the past weather to predict the future weather, the input and output are both weather); in classification, the input and output are in different semantic space (e.g., using ECG to predict heart arrhythmia, the input ECG signal and output cardiovascular are different). This work focuses on time series classification, investigating contrastive and generative learning methods instead of autoregressive approaches.

Self-supervised \textit{contrastive} learning was originally proposed in SimCLR~\cite{chen2020simple} for image processing. It was then rapidly adapted to the time series domain and has shown remarkable success~\cite{yang2022timeclr}. It aims to learn representations by maximizing the similarity between positive pairs (e.g., two augmented views of the same sample) while minimizing the similarity between negative pairs (e.g., two different samples)~\cite{chen2020simple}. Recent works, such as TS2Vec~\cite{yue2022ts2vec} and TF-C~\cite{zhangself}, have demonstrated the effectiveness of contrastive learning for time series representation learning. However, contrastive learning has a limited understanding of data distribution. The core idea behind contrastive learning is sample discrimination, which means distinguishing each sample from others, instead of understanding the underlying data distribution, which limits its ability to capture higher-level pictures of the whole dataset.

On the other hand, generative learning approaches, such as masked autoencoders (MAE)~\cite{he2022masked}, focus on learning the underlying data distribution by reconstructing or generating realistic time series samples. MAEs randomly mask a portion of the input data and train the model to reconstruct the masked regions, encouraging the model to learn robust and generalizable representations~\cite{feichtenhofer2022masked}. Generative methods have also been expanded to time series analysis tailored for temporal feature learning, such as TS-MAE~\cite{zerveas2021tsmae}, MTS-MAE~\cite{tang2022mtsmae}, and TimeMAE~\cite{cheng2023timemae}. However, purely generative methods focus on data generation or reconstruction rather than feature discrimination, thus often lacking the discriminative power needed for downstream classification tasks. 

\begin{figure*}
    \centering
    \includegraphics[width=\linewidth]{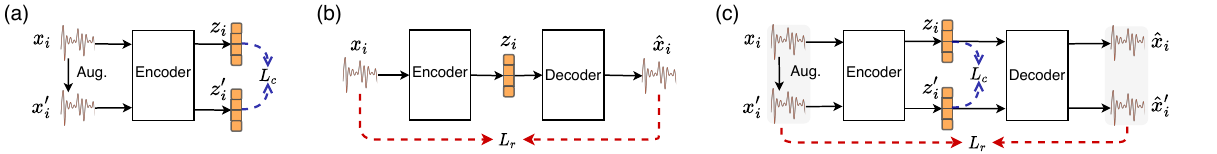}
    \caption{Frameworks of self-supervised learning approaches. (a) Contrastive SSL (e.g., SimCLR) focuses on sample-level discrimination while having a limited understanding of the distribution of the entire dataset. \textit{Aug.} denotes augmentation. (b) Generative SSL (e.g., MAE) aims to learn reconstructive representations but omits the discriminative power. (c) Proposed \name\ unifying the advantages of contrastive and generative methods. 
    }
    \label{fig:keyidea}
\end{figure*}

In this work, we propose a novel framework, \textbf{Co}ntrastive \textbf{Gen}erative framework for \textbf{T}ime series (\name),
% \textbf{Ma}sked \textbf{Co}ntrastive autoencoder (MaCo), 
to unify contrastive and generative SSL methods for mutual benefits. Figure~\ref{fig:keyidea} illustrates how \name\ is different from a standard contrastive method along with a generative model.
\name\ is simple but effective: upon the backbone of MAE, augmenting the samples in the input space, and calculating the contrastive loss in the embedding space. In specific, we feed time series samples (along with the augmented views) into an encoder for latent feature learning, then calculate contrastive loss in the embedding space, and then decode the embeddings to reconstruct the input sample. 
We evaluated \name\ on 6 diverse datasets ranging from mechanical fault detection, heart disease, electrical usage behavior to human activity recognition, covering broad time series channels (1 to 12), sample length (96 to 1280), classes (binary to 7-class), and frequencies (from one observation per five minutes to 16 KHz). 
Experimental results support \name's effectiveness: 
% \name\ outperforms the non-pretraining method by up to 96.5\% in F1 (relative margin), and also 
this unified framework outperforms the sole contrastive pre-training and generative pre-training up to 59.2\% and 14.3\% in F1 score (FD dataset), respectively.

The key contributions of this work include:
\begin{itemize}
    \item We propose a unified framework, \name, to leverage contrastive and generative SSL methods, to capture both the high-level dataset distributions and fine-grained sample discrimination. \name\ learns representations with discriminative power for time series classification.
    \item We evaluate \name\ on 6 diverse time series datasets. The results show the proposed unified \name\ consistently outperforms standalone contrastive and generative methods, achieving state-of-the-art results. Our unified approach bridges the gap between global data modeling and instance-level discrimination.
\end{itemize}

\section{Related Work}
\label{sec:related_work}

\subsection{Self-Supervised Contrastive Learning in Time Series} 
Contrastive learning has emerged as a dominant paradigm in unsupervised representation learning, initially gaining momentum in the vision domain. A seminal contribution in this area is SimCLR which demonstrated that a simple contrastive framework could yield state-of-the-art results across multiple visual benchmarks by maximizing agreement between augmented views of the same sample using a contrastive loss~\cite{chen2020simple}. Although originally tailored for image data, SimCLR inspired adaptations to time series, catalyzing a surge of interest in contrastive learning techniques within this domain.

One of the early and influential efforts in this transition was Contrastive Predictive Coding (CPC) by Franceschi et al., which applied autoregressive models to learn representations from multivariate time series, significantly improving classification performance~\cite{henaff2020data}. Expanding on the temporal structure of time series, Zhang et al. proposed the Time-Frequency Consistency (TF-C) model, which aligns representations in both time and frequency domains, thereby enhancing the robustness and accuracy of learned features~\cite{zhangself}.

More targeted approaches for time series have continued to evolve. Eldele et al. introduced TS-TCC, a temporal contrastive learning framework that explicitly leverages sequential dependencies to learn rich and generalizable representations~\cite{eldele2021time}. Woo et al. proposed CoST, a dual-domain method that combines contrastive learning in both temporal and spectral spaces to improve both forecasting and classification tasks~\cite{woocost2022}. 
Yue et al. advanced the field further with TS2Vec, a universal time series representation model utilizing a hierarchical contrastive loss to capture patterns at multiple temporal resolutions~\cite{yue2022ts2vec}. 
Recently, Lee et al. proposed SoftCLT, incorporating soft assignments in contrastive losses to better capture intra-sample correlations~\cite{leesoft2022}. 
Shamba et al. designed DynaCL, which uses temporal adjacency and N-pair loss to dynamically determine positive pairs, improving robustness across datasets~\cite{shamba2024dynamic}. 
% Peng et al. presented CDCC, a cross-domain model that aligns time-frequency representations through instance and cluster-level constraints~\cite{peng2024cdcc}. 
These works reflect a rapidly evolving landscape in contrastive representation learning for time series classification.

\subsection{Self-Supervised Generative Learning in Time Series}
Generative self-supervised learning is another foundational paradigm in deep learning for unsupervised representation learning. Generative SSL aims to reconstruct or generate new data samples from learned representations, with most early work grounded in autoencoder and Generative Adversarial Network (GAN) architectures~\cite{li2023timae,liang2022self}.

% One of the pioneering methods in this space is the Variational Autoencoder (VAE)\cite{kingmaauto}, which introduces a probabilistic latent space for reconstructing inputs and generating new data samples. VAEs have proven effective in modeling complex data distributions and are frequently used in time series synthesis and anomaly detection. Similarly, Generative Adversarial Networks (GANs)\cite{goodfellow2020generative} have played a central role in generative SSL. These models consist of a generator-discriminator pair trained in an adversarial setting, enabling the generation of realistic and diverse samples.  

Beyond traditional architectures, more structured generative models have emerged. The development of the Masked Autoencoder (MAE)~\cite{he2022masked} marked another significant milestone. By masking portions of input data and training the model to reconstruct the masked values, MAEs enable efficient representation learning. This strategy has been adapted into time series data, such as MTSMAE~\cite{tang2022mtsmae}, TS-MAE~\cite{zerveas2021tsmae}, TFMAE~\cite{fang2024tfmae}, Ti-MAE~\cite{li2023timae}, HSCMAE~\cite{gao2023hscmae}, MAEEG~\cite{zhang2022maeeeg}, STMAE~\cite{jiang2023stmae}, and SimMTM~\cite{dong2023simmtm}. 
Specifically, Li et al. introduced Ti-MAE, a self-supervised approach that randomly masks embedded time series data and trains a Transformer-based autoencoder to reconstruct them at the point level. This technique bridges generative modeling and representation learning, improving both forecasting and classification accuracy~\cite{li2023timae}.
Cheng et al. proposed TimeMAE, which applies a window-slicing strategy followed by localized random masking of sub-series. By using a decoupled encoder-decoder architecture, TimeMAE efficiently captures temporal dependencies for effective time series representation learning~\cite{cheng2023timemae}.
Zha et al. developed ExtraMAE, a scalable MAE-based framework for time series generation. By recovering masked patches of time series data, ExtraMAE learns temporal dynamics and achieves strong performance in classification, imputation, and prediction tasks~\cite{zha2022extramae}.
Fang et al. introduced TFMAE, which combines temporal and frequency domain masking strategies using Transformer-based autoencoders. This dual masking approach improves robustness in detecting time series anomalies under distributional shifts~\cite{fang2024tfmae}.
Tang and Zhang presented MTSMAE, a masked autoencoder model for multivariate time-series forecasting. It uses a novel patch embedding strategy tailored to time-series structure, enabling improved forecasting performance compared to standard Transformer-based models~\cite{tang2022mtsmae}.

\subsection{Combining Contrastive and Generative Learning} 
% 1. Contrastive and Generative Graph Convolutional Networks for Graph-based Semi-Supervised Learning

% 2. Contrastive Generative Self-Supervised Learning for 3D Medical Image Segmentation

% 3. Generative and contrastive based self-supervised learning model for histop

% 4. Contrastive Masked Autoencoders are Stronger Vision Learners

% 5. CAN: A simple, efficient and scalable contrastive masked autoencoder framework for learning visual representations

% 6. HiCMAE: Hierarchical Contrastive Masked Autoencoder for self-supervised Audio-Visual Emotion Recognition

To the writing date of this manuscript, we found 5 publications in SSL that have explored hybrid contrastive-generative approaches, but mainly focusing on image~\cite {huang2023contrastive,li2021generative,zhang2021contrastive,chen2022can} and graph~\cite{wan2020contrastive}. However, they face key limitations when applied to \emph{time series classification}. Graph-based methods lack temporal modeling; vision-centric MAE uses spatially random masking, unsuitable for sequential data; and audio-visual hybrids (e.g., \cite{wang2022hicmae}) prioritize cross-modal alignment over intra-modal temporal dynamics. Crucially, no existing work jointly optimizes contrastive and generative objectives for time series, leaving a gap in temporal-aware representation learning. In this work, we propose a novel framework that bridges the divide between contrastive and generative SSL, offering an effective solution for time series classification.
% where both invariance and temporal fidelity are essential. 

\section{Method}

\begin{figure*}
    \centering
    \includegraphics[trim=10 0 8 0, clip, width=\linewidth]{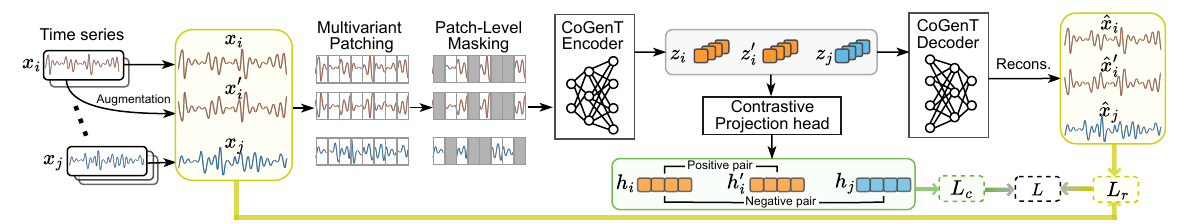}
    \caption{Framework of the proposed \name. 
    \name\ can directly work on multivariate time series while the sample $\bm{x}_i$ is illustrated as univariate for simplicity.
    The $\bm{z}_i$ is the learned representation for $\bm{x}_i$. We add a contrastive project head to map it to $\bm{h}_i$ for contrastive loss calculation following the setting of SimCLR~\cite{chen2020simple} while also supported by our preliminary experiments. 
    } 
    \label{fig:framework}
\end{figure*}

\subsection{Overview}

% \textbf{Notation and Problem Formulation.}
Given a multivariate time series sample $\bm{x}_i \in \mathbb{R}^{T \times D}$, where $T$ denotes the sequence length (or time steps) and $D$ denotes the number of channels. 
We aim to learn a function $f: \bm{x}_i \rightarrow \bm{z}_i$ where $\bm{z}_i$ is a high-dimensional representation of time series $\bm{x}_i$. The learned $\bm{z}_i$ will be fed into a downstream classifier $g$ for classification. 

The model learning contains two stages: self-supervised pre-training to learn $f$ and a supervised fine-tuning to learn $g$. The focus of this work is on the self-supervised learning of $f$.

\subsection{Model Architecture}
The proposed \name, as shown in Figure~\ref{fig:framework}, contains several key modules: augmentation, multivariate time series patching, patch-level masking, \name\ encoder, \name\ decoder, and a contrastive projection head.  
% \xiang{overview, novelty}

\header{Time Series Augmentation} We augment each time series sample $\bm{x}_i$ to generate a new view $\bm{x}'_i$. These augmented samples facilitate the construction of positive and negative pairs, which are critical for contrastive learning. By optimizing a contrastive loss function, leveraging pairwise similarities and dissimilarities, the model learns discriminative representations without reliance on ground truth.

% The augmented samples could construct positive and negative pairs in representation learning, further enabling contrastive loss that enables model optimization without requiring ground truth; this is how the self-supervised learning is empowered. 

% so that the contrastive distance can be measured. Different from a standard contrastive learning framework, we augment the patches instead of the whole sample. 
Among the eight popular augmentation methods in time series~\cite{liu2023self}, we select the most suitable ones following the augmentation selection guidelines based on the signals' seasonality and trend~\cite{liu2024guidelines}. 
Taking jittering augmentation as an example, the augmented $\bm{x}'_i = \bm{x}_i + \epsilon \mathcal{N} (0, 1)$ means adding random Gaussian noise on the original sample, where $\epsilon$ is a coefficient to adjust the amplitude of noise. 

\header{Multivariate Time Series Patching}
Following PatchTST~\cite{nie2022time} and MAE~\cite{he2022masked}, we partition the input time series \(\bm{x}_i\) into non-overlapping patches to enhance local pattern recognition and computational efficiency. This patching strategy enables the model to hierarchically process short-term dependencies (e.g., transient spikes or periodic segments) before inferring global temporal dynamics.  

The patch size \([L, D]\) is a critical hyperparameter. In which the \(L\) denotes the number of consecutive time steps per patch, controlling the granularity of local feature extraction, and the \(D\) represents the full sensor dimension, ensuring each patch retains multivariate channel information~\cite{wang2024medformer}.  

With the given patch size, we split $\bm{x}_i \in \mathbb{R}^{T \times D}$ into $N$ patches where $N = \lfloor T / L \rfloor$. The patched sequence is:
\begin{equation}
\bm{x}_i^{\text{patched}} 
% = [\bm{p}_{i, 1}, \dots, \bm{p}_{i, N}] \in \mathbb{R}^{N \times L \times D},
=\begin{bmatrix}
\bm{p}_{i, 1} \\
\bm{p}_{i, 2} \\
\cdots\\
\bm{p}_{i, N}
\end{bmatrix}
\in \mathbb{R}^{N \times L \times D}
\end{equation} 
Each patch $\bm{p}_{i, 1} \in \mathbb{R}^{L \times D}$ represents a local temporal window. 
In this paper, a notation with a single subscript denotes \textit{sample-level variables}, while a notation with two subscripts indicates \textit{patch-level variables}.

% \subsection{Patch-Level Masking Module}
\header{Patch-Level Masking}
This module randomly drops out some patches, only keeping partial information, forcing the model to learn robust representations for reconstruction~\cite{cheng2023timemae}. The dropout is achieved by binary masking. 
  
A binary mask $\bm{m} \in \{0,1\}^{N}$ is sampled with masking ratio $\theta$. 
% High masking ratios 
% (e.g., $p=0.8$) 
% prevent trivial solutions and encourage learning of cross-patch dependencies. 
The mask is a vector with $N$ elements. After masking, the remaining sample can be written as
\begin{equation}
% $$
\tilde{\bm{x}}_i^{\text{patched}} = \bm{x}_i^{\text{patched}} \circ \bm{m}
% $$   
\end{equation}
where $\circ$ denotes element-wise multiplication. After masking, there are $N*(1-\theta)$ out to $N$ patches remaining. Following the setups in MAE, PatchTST, along with our experimental hyper-parameter tuning, we set $\theta=0.75$ in this work.
Note that the above patching and masking apply to all samples and augmented samples independently.
% For simplicity, the elaboration below will be at the patch level. 

\header{\name\ Encoder}
The encoder maps unmasked patches into latent representations, hierarchically aggregating temporal features. The encoder distills each patch into a representation while preserving temporal relationships. 

We feed the patched sample $\tilde{\bm{x}_i}^{\text{patched}}$ into the Encoder, in other words, feed all the patches belonging to the sample into the encoder, for the learning of representation $\bm{z}_i$: 
\begin{equation}
    \bm{z}_i = \text{Encoder}(\tilde{\bm{x}_i}^{\text{patched}}),
\end{equation}
or, at the patch level,
\begin{equation}
\bm{z}_i =
\begin{bmatrix}
\bm{z}_{i, 1} \\
\bm{z}_{i, 2} \\
\cdots\\
\bm{z}_{i, N}
\end{bmatrix}
= \text{Encoder}( \begin{bmatrix}
\bm{p}_{i, 1} \\
\bm{p}_{i, 2} \\
\cdots\\
\bm{p}_{i, N}
\end{bmatrix})
\end{equation}
The learned $\bm{z}_{i, 1}$ is the representation of patch $\bm{p}_{i, 1}$. Similarly, this operation applies to a patch $\bm{p}'_i $ comes from the augmented sample $\bm{x}'_i$ and generates corresponding representation $\bm{z}'_{i, 1}$.

% We consider a patch $\bm{p}_i \in \tilde{\bm{x}_i}^{\text{patched}}$, the corresponding patch c from  augmented sample $\bm{x}'_i$, and a patch $\bm{p}_j$ comes from a different sample $\bm{x}_j $. After mapping, we get the representations:
% \begin{equation}
%   % $$
%   \bm{z}_i = \text{Encoder}(\bm{p}_i), \quad \bm{z}'_i = \text{Encoder}(\bm{p}'_i), \quad \bm{z}_j = \text{Encoder}(\bm{p}_j)
%   % $$  
%   \vspace{1mm}
% \end{equation}

In this work, guided by our preliminary experiments, we adopt a 2-layer Transformer as the encoder backbone. The encoder architecture $\text{Encoder}(\cdot)$ comprises three key components~\cite{he2022masked}: a) a linear projection layer that maps input patches to a higher-dimensional space, enhancing the model's representational capacity;
b) positional encoding that injects temporal order information into the patch representations;
c) and two stacked Transformer blocks.
% to capture both local and global temporal dependencies.

\header{Contrastive Projection Head}
Here is one key difference with a standard MAE.
We measure the contrastive relationship of the learned representations. 
We can regard the contrastive loss as a regularization that encourages the model to learn similar representations for similar input samples (or positive samples), vice versa. 

Here we regard the augmented sample and the original sample as a positive pair such as $(\bm{x}_i, \bm{x}'_i)$ while two different samples as a negative pair such as $(\bm{x}_i, \bm{x}_j)$. 

We further employ a projection head to map the encoded representation $\bm{z}_i$ to another embedding $\bm{h}_i$:  
\begin{equation}
  \bm{h}_i = \text{ProjectHead} (\bm{z}_i).
\end{equation}
To make it clear, we also provide the patch-level equation:
\begin{equation}
  \bm{h}_i = \text{ProjectHead} ([\bm{z}_{i, 1}|| \bm{z}_{i,2 }|| \cdots || \bm{z}_{i, N}]),
\end{equation}
where $\cdot||\cdot$ denotes concatenation or flatten. 
As demonstrated in SimCLR, also verified by our preliminary empirical results, the loss function calculated on $\bm{h}_i$ achieves better downstream classification performance than on $\bm{z}_i$. In this work, the project head is implemented as one fully-connected layer. 

Please note that $\bm{h}_i$ is only used for contrastive loss calculation, while reconstructions are derived from $\bm{z}_i$. In other words, the representation fed into the downstream classifier is $\bm{z}_i$. Generally, the representation $\bm{z}_i$ has higher dimension that sample $\bm{x}_i$ to get higher expression power; the embedding $\bm{h}_i$ has lower dimension than $\bm{z}_i$ to condense the information for easier contrastive distance calculation. 
The rationale for using $\bm{h}_i$ (instead of $\bm{z}_i$) for contrastive loss follows the standard pipeline of SimCLR~\cite{chen2020simple}, a convention adopted by nearly all contrastive SSL. 
While we adhere to this established setup, a key distinction arises in our framework: 1) in typical contrastive SSL, $\bm{z}$ represents a single embedding vector (rank-1) for an input sample, while in \name, $\bm{z}$ denotes a set of patch embeddings (rank-2). In this work, we simply concatenate all the patch embeddings and feed them into the project head; however, more technical fusing methods (such as patch-wise attention mechanism) could be explored in the future. 

% Please note, the $\bm{h}_i$ is only used for contrastive loss calculation while the reconstruction still comes from the $\bm{z}_i$. In other words, the representation that will be fed into the downstream classifier is $\bm{z}_i$. The reason of using instead of h for contrastive loss is provided in following the pipeline of SimSLR~\cite{chen2020simple} and is widely followed in almost all contrastive SSL. We here simply followup the popular setup, but the difference is that: in general contrastive SSL, the z denotes a single embedding (rank of 1) of input sample, but here in \name, z denotes a set of patch embeddings with rank of 2.  

\header{\name\  Decoder} 
This decoder reconstructs the patch from learned representation $\bm{z}_i$. The architecture of this decoder is identical to the MAE decoder. We call the reconstructed patch as $\bar{\bm{p}}_i$:
\begin{equation}
    % $$
    \bar{\bm{p}}_i = \text{Decoder}(\bm{z}_i)
    % .$$
\end{equation}

\subsection{Loss Function}
To achieve self-supervised learning, we need a loss function that can be used to optimize the model through back propagation while the ground truth is unavailable. To obtain the loss function, we consider two objectives: the loss in reconstructing the patch and the contrastive loss between positive and negative pairs.

\header{Reconstruction Loss}
We adapt the widely used Mean Squared Error (MSE) to measure the Euclidean distance between the original patches and reconstructed patches. The loss calculation only considers the unmasked patches. Different from a standard MAE model, we have both the original sample $x_i$ and the augmented sample $x'_i$, both of which pass through the patching, masking, encoder, and decoder process, so that we can measure the reconstruction loss $\mathcal{L}_{r}$ for both of them:
\begin{equation}
    % $$
    \mathcal{L}_{r} = \frac{1}{|\mathcal{M}|} \sum_{i \in \mathcal{M}} (\|\hat{\bm{p}}_i - \bm{p}_i\|^2+
\|\hat{\bm{p}'}_i - \bm{p}'_i\|^2)
% ,$$
\end{equation}

where $\mathcal{M}$ is the set of unmasked patches. For example, in a batch with $B$ samples, each sample has $N*(1-\theta)$ unmasked patches, there are $B*N*(1-\theta)$ patches in total: $|\mathcal{M}| = B*N*(1-\theta)$.
When calculating $\mathcal{L}_{r}$, we consider not only the original samples but also the augmented samples. 

\header{Contrastive Loss}
We apply the Normalized Temperature-Scaled Cross-Entropy (NT-Xent) as contrastive loss to minimize distance between positives while maximizing separation from negatives:
% \begin{equation}
%     % \[
%     \mathcal{L}_{c} = - \frac{1}{2|\mathcal{M}|} \sum_{i \in \mathcal{M}} (log \frac{\exp(\text{sim}(\bm{h}_i, 
%     \bm{h}'_i)/\tau)}{\sum_{k \in \mathcal{M}} \exp(\text{sim}(\bm{h}_i, \bm{h}_k)/\tau)}),
%     % \]
% \end{equation}
\begin{equation}
    \mathcal{L}_{c} = - \frac{1}{B} \sum_{i=1}^B (log \frac{\exp(\text{sim}(\bm{h}_i, 
    \bm{h}'_i)/\tau)}{\sum_{k =1}^{2B} \exp(\text{sim}(\bm{h}_i, \bm{h}_k)/\tau)}),
\end{equation}
where $\text{sim}$ is cosine similarity, the $\tau$ is a temperature hyperrparameter for normalization. 
% There are two differences between this loss function and a standard contrastive loss. First, our loss is calculated on patch levels instead of sample levels. Second, the negative set is not only the different samples, it also includes all the patches that come from a different position from $\bm{p}_i$. For each patch $\bm{p}_i$, there will be only one positive pair $(\bm{p}_i, \bm{p}'_i)$, but $2|\mathcal{M}| -1$ negative pairs. 

\header{Joint Loss}
To enable end-to-end training of the unified framework, we formulate a composite loss function $\mathcal{L}$ that jointly optimizes the reconstruction and contrastive losses:
\begin{equation}
\label{eq:total_loss}
    \mathcal{L} = \lambda_c \mathcal{L}_{c} + \lambda_r \mathcal{L}_{r}.
\end{equation}
Here, the $\lambda_c$ and $\lambda_r$ are hyperparameters to adjust the scale of the two loss items.

\section{Experiments} 

\subsection{Datasets}
\textbf{FD.} The Faulty Detection (FD) dataset ~\cite{lessmeier2016condition} includes bearing condition monitoring data from electromechanical drive systems. The signals are classified into three types based on the bearing's condition: undamaged, inner damage, and outer damage.
Following ~\cite{zhang2022self}, we employ preprocessed data with a reduced window length of 1,280 (down from 5,120) to enhance computational efficiency. 

\textbf{FordA.} This dataset~\cite{ucrarchive} comprises automotive engine noise recordings for binary classification of \textit{faulty} versus \textit{normal} components. This dataset is widely used to evaluate methods for detecting high-frequency mechanical anomalies. 
It originates from the IEEE World Congress on Computational Intelligence held in 2008. While the dataset remains widely used for benchmarking in detecting high-frequency mechanical anomalies, the original competition website has since been deactivated, rendering metadata such as sampling frequency irretrievable. While recording engine noise typically requires sampling rates of 48 kHz or 96 kHz for optimal fidelity, we conservatively estimate that the FordA dataset was originally sampled at larger than 10 KHz. 

\textbf{ClCon.} The Chlorine Concentration (ClCon) dataset~\cite{li2011fast,ucrarchive} contains simulated water quality measurements. It records chlorine concentration levels at 166 pipe junctions sampled every 5 minutes over 15 days (4,310 timesteps), capturing system-wide chemical dynamics. The target is to classify chlorine concentration levels into \textit{high} or \textit{low}. 

\textbf{PTB.} Physikalisch-Technische Bundesanstalt (PTB) dataset measures 12-lead ECG signals~\cite{bousseljot1995nutzung, goldberger2000physiobank}. We make it a binary task: distinguish Myocardial infarction from Healthy. We employ subject-independent data split, which means the training set and testing set don't have overlapped subjects~\cite{wang2024evaluate}.
% Subject IDs and data splits will be released for reproducibility.  

\textbf{ElecD.} It measures household electricity usage, which is sampled from 251 households at two-minute intervals for a month~\cite{lines2011classification}. The 7 classes correspond to 7 kinds of consumers' electricity usage behavior. As the original dataset doesn't have a validation set, we regard the test set as the validation set.  

\textbf{HAR.} Human Activity Recognition (HAR) dataset~\cite{misc_human_activity_recognition_using_smartphones_240} contains recordings of 30 health subjects performing six kinds of daily tasks: waking, walking upstairs, walking downstairs, sitting, standing, lying. 
A smartphone records 3-axial linear acceleration during the tasks at 50Hz. We follow ~\cite{zhang2022self} for the data split of training, validation, and test sets.  

All datasets are evaluated using the following procedure. During the pretraining stage, we randomly split the training set into two subsets: 90\% is used for self-supervised pretraining, while the remaining 10\% is reserved for sanity checks to monitor model convergence.

The pretrained model is then transferred to the fine-tuning stage. Here, we randomly sample 30\% of the training set (a label ratio of 0.3, consistent with common practices~\cite{liu2023self}) for supervised fine-tuning. The validation set is used to select optimal hyperparameters and model configurations, and final performance is assessed on the held-out test set.

\begin{table}[]
\label{tab:datasets}
\setlength{\tabcolsep}{2pt} % Reduce column padding
\centering
\caption{Dataset statistics}
\resizebox{0.5\textwidth}{!}{
\begin{tabular}{@{}cccccccccc@{}}
\toprule
\multirow{2}{*}{\textbf{Dataset}} & \multirow{2}{*}{\textbf{Class}} & \multirow{2}{*}{\textbf{Channel}} & \multirow{2}{*}{\textbf{Length}} & \multicolumn{2}{c}{\textbf{Pre-train datasize}} & \multicolumn{3}{c}{\textbf{Fine-tune datasize}} & \multirow{2}{*}{\textbf{Frequency}} \\ \cmidrule(lr){5-9}
 &  &  &  & \textbf{Train} & \textbf{Val} & \textbf{Train} & \textbf{Val} & \textbf{Test} &  \\ \midrule
\textbf{FD} & 3 & 1 & 1280 & 2008 & 224 & 669 & 2728 & 2728 & 16 KHz \\
\textbf{FordA} & 2 & 1 & 500 & 2592 & 288 & 863 & 721 & 1320 & \textbackslash \\
\textbf{ClCon} & 3 & 1 & 166 & 336 & 37 & 110 & 94 & 3840 & per 5 mins \\
\textbf{PTB} & 2 & 12 & 600 & 22257 & 2473 & 7418 & 3020 & 3365 & 250 Hz \\
\textbf{ElecD} & 7 & 1 & 96 & 8033 & 893 & 2675 & 7711 & 7711 & per 2 mins \\
\textbf{HAR} & 6 & 3 & 206 & 5293 & 588 & 1762 & 1471 & 2947 & 50 Hz \\ \bottomrule
\end{tabular}
}
\vspace{-3mm}
\end{table}

% \begin{table*}[]
% \caption{}
% \label{tab:FD}
% \begin{tabular}{@{}ccccccccccc@{}}
% \toprule
% \multirow{2}{*}{\textbf{Dataset}} & \multirow{2}{*}{\textbf{\# class}} & \multirow{2}{*}{\textbf{\# channel}} & \multirow{2}{*}{\textbf{ts length}} & \multicolumn{2}{c}{\textbf{Pre-train datasize}} & \multicolumn{3}{c}{\textbf{Fine-tune datasize}} & \multirow{2}{*}{\textbf{Frequency}} & \multirow{2}{*}{\textbf{Dataset description}} \\ \cmidrule(lr){5-9}
%  &  &  &  & \textbf{Train} & \textbf{Val} & \textbf{Train} & \textbf{Val} & \textbf{Test} &  &  \\ \midrule
% \textbf{FD} & 3 & 1 & \begin{tabular}[c]{@{}c@{}}1280\\ (downsampled)\end{tabular} & 2008 & 224 & 669 & 2728 & 2728 & 16 KHz & \begin{tabular}[c]{@{}c@{}}Faulty Detection in \\ electromechanical drive system\end{tabular} \\
% \textbf{FordA} & 2 & 1 & 500 & 2592 & 288 & 863 & 721 & 1320 & \textbackslash{} & \begin{tabular}[c]{@{}c@{}}Engine noise based \\ automotive diagnostics\end{tabular} \\
% \textbf{ClCon} & 3 & 1 & 166 & 336 & 37 & 110 & 94 & 3840 & per 5 mins & \begin{tabular}[c]{@{}c@{}}Simulated chlorine control \\ in water treatment\end{tabular} \\
% \textbf{PTB} & 2 & 12 & \begin{tabular}[c]{@{}c@{}}600\\ (resegmented)\end{tabular} & 22257 & 2473 & 7418 & 3020 & 3365 & 250 Hz & Heart disease diagnosis (ECG) \\
% \textbf{ElecD} & 7 & 1 & 96 & 8033 & 893 & 2675 & 7711 & 7711 & per 2 mins & \begin{tabular}[c]{@{}c@{}}Electronic device \\ usage monitoring\end{tabular} \\
% \textbf{HAR} & 6 & 3 & 206 & 5293 & 5
% 88 & 1762 & 1471 & 2947 & 50 Hz & Human activity recognition \\ \bottomrule
% \end{tabular}
% \end{table*}

\subsection{Experimental Setup}
% \header{Baselines.}
\subsubsection{Baselines}
We select SimCLR and MAE as representative baselines for contrastive and generative self-supervised learning, respectively, as they constitute the most fundamental and widely adopted frameworks in their respective paradigms.

While numerous advanced variants of SimCLR and MAE exist, we intentionally compare against their vanilla implementations for two important reasons: proof-of-concept and extensibility. First, by using the original SimCLR and MAE architectures without modifications, we can isolate and demonstrate the fundamental advantages of combining contrastive and generative learning in our unified framework. This conservative comparison ensures any performance improvements are attributable to our methodological innovation rather than architectural refinements.
Second, \name\ has strong extensibility. The modular design of \name\ means that future work can readily substitute these baseline components with more advanced contrastive (e.g., TS2Vec) or generative (e.g., Ti-MAE) architectures as described in Section~\ref{sec:related_work}. If \name\ outperforms vanilla SimCLR and MAE, it follows logically that upgraded versions would maintain this advantage over their standalone counterparts.

% While variants of these methods exist, we use their vanilla forms to isolate the benefits of unifying contrastive and generative learning in \name. Outperforming these baselines validates that \name's hybrid design is inherently superior, irrespective of architectural upgrades (e.g., replacing SimCLR/MAE with TS2Vec/Ti-MAE; see Section~\ref{sec:related_work}). 

We adapt SimCLR and MAE from image to time series processing by treating multivariate time series as grey-scale (1-depth) images while retaining their original architectures. All the settings follow the original papers except where explicitly stated. Hyperparameters are adjusted to align with the temporal dimensionality of the input data. 
% To ensure full reproducibility, we will release all implementation details and code. 

For a comprehensive evaluation, we consider \name\ with pretrained SimCLR (contrastive only), pretrained MAE (generative only), and their non-pretrained versions. The non-pretraining model means training the model from scratch, only optimizing the classification cross-entropy loss while excluding all self-supervised losses, no matter whether contrastive or generative. In such a case, the MAE and \name\ are identical. 
% All the preprocessed datasets will be released upon acceptance.

% ElecD: 4.14 million parameters
% FD: 18,199,616 (18.2 million)
\header{Implementation Details} 
% \xiang{Encoder structure, hyper-parameters
% set up. label ratio = 0.3 }
% \ziyu{add joint loss hyperparameter details}\\
The implementation details and code will be made publicly available in the GitHub repository to ensure full reproducibility of the results.
Here we report the essential experimental details, taking FD as an example. 
We chose time-masking (randomly mask out 50\% of timepoints in the sample) as augmentation. 
% (Normal Gaussian noise; \xiang{0.1} coefficient in jittering addition) 
We set the patch size as $[1, 64]$, leading to 20 patches. The masking ratio is 0.75, resulting in 5 unmasked patches, along with one global representation (i.e., classification token) that aggregates information from all the patches. So the input to the Encoder is a matrix with 6 rows and 64 columns. 

For easier comparison, we use the same encoder backbone as in MAE~\cite{he2022masked}. In detail,
first, the input is mapped into a latent representation with 512 dimensions via a linear projector with ReLU activation;
then the encoder includes a patch embedding layer using a 1D convolution that transforms the input sequence into patch tokens with 512 dimensions, followed by a Transformer composed of two attention blocks. 
Each block consists of multi-head self-attention and a feedforward MLP with GELU activation, all normalized by LayerNorm. 
We set the hyperparameters $\lambda_c$ and $\lambda_r$ to balance the scale of $\mathcal{L}_{c}$ and $\mathcal{L}_r$ to make $\lambda_c \mathcal{L}_c$ roughly equals to $\lambda_r \mathcal{L}_r$ in the first epoch.

The decoder mirrors the encoder's structure, using two Transformer blocks and a linear head layer to reconstruct the masked patches. A Rearrange operation reshapes the decoded patches back into the original signal format. Notably, the entire architecture maintains a compact footprint of approximately 18.2 million parameters, enabling efficient training and inference while preserving strong representational capacity for time-series data reconstruction and downstream tasks.

The contrastive projection head is a simple 2-layer fully connected architecture. The downstream classifier in the fine-tuning stage is also a 2-layer MLP. In which, the dimension in the hidden layer is 10\% of the representation dimension $\bm{z}_i$. 
% \xiang{Check the dimension of h}

\header{Computational resources} Taking FD as an example, the model size for SimCLR, MAE, and \name\ are 16.9M, 12.7M, and 18.2M parameters, respectively. On one hand, although 18.2 million parameters is small compared with CV and NLP models, it's large in the time series area. On the other hand, \name\ is at the same level as existing widely-used SSL frameworks, not causing extra computational burden.
The experiments in this work are conducted on a workstation equipped with an NVIDIA RTX A4000 GPU card featuring 16 GB of memory, and an Intel Core i9 CPU with 32 GB of RAM.
The pretraining time for 100 epochs on FD is 175.08s. 

% \begin{table}[]
% \caption{}
% \label{tab:my-table}
% \begin{tabular}{cccc}
% \textbf{parameter} & \textbf{SimCLR} & \textbf{MAE} & \textbf{CoGenT} \\
% \textbf{FD} & 16,900,736 & 12,692,544 & 18,199,616 \\
% \multicolumn{1}{l}{} & \multicolumn{1}{l}{} & \multicolumn{1}{l}{} & \multicolumn{1}{l}{}
% \end{tabular}
% \end{table}

\subsection{Experimental Results and Analysis}

\begin{table*}[h]
\centering
\caption{Results comparison on 6 time series datasets. The reported metric is the F1 score. The `w/o' and `w/' denote `without' and `with', respectively. MAE and \name\ have the same architecture in the no-pretraining setup, so we merged them into one row.}
\label{tab:results}
\begin{tabular}{@{}clcccccc@{}}
\toprule
\textbf{Pretain} & \textbf{Method} & \textbf{FD} & \textbf{FordA} & \textbf{ClCon} & \textbf{PTB} & \textbf{ElecD} & \textbf{HAR} \\ \midrule
 \multirow{2}{*}{\textbf{w/o}} & \textbf{SimCLR} & 0.5460 $\pm$ 0.0141 & 0.7244 $\pm$ 0.0052 & 0.3194 $\pm$ 0.1057 & 0.5384 $\pm$ 0.0162 & 0.5301 $\pm$ 0.0091 & 0.8757 $\pm$ 0.0022 \\
 & \textbf{MAE/\name} & 0.8894 $\pm$ 0.0117 & 0.8993 $\pm$ 0.0065 & 0.3862 $\pm$ 0.0206 & 0.5971 $\pm$ 0.0179 & 0.5647 $\pm$ 0.0130 & 0.8889 $\pm$ 0.0202 \\ \midrule
 \multirow{3}{*}{\textbf{w/}}
 & \textbf{SimCLR} & 0.6063 $\pm$ 0.0183 & 0.6741 $\pm$ 0.0107 & 0.3282 $\pm$ 0.1003 & 0.5495 $\pm$ 0.0234 & 0.5515 $\pm$ 0.0065 & 0.8931 $\pm$ 0.0062 \\
 & \textbf{MAE} & 0.8447 $\pm$ 0.0045 & 0.8828 $\pm$ 0.0030 & 0.3889 $\pm$ 0.0106 & 0.5917 $\pm$ 0.0067 & 0.5663 $\pm$ 0.0085 & 0.8758 $\pm$ 0.0094 \\
 & \textbf{\name} & \textbf{0.9652 $\pm$ 0.0039} & \textbf{0.9131 $\pm$ 0.0034} & \textbf{0.4202 $\pm$ 0.0090} & \textbf{0.6032 $\pm$ 0.0164} & \textbf{0.5948 $\pm$ 0.0031} & \textbf{0.8965 $\pm$ 0.0074} \\ \midrule
\multicolumn{2}{c}{\textbf{Boost over SimCLR}} & 59.20\% & 35.45\% & 28.03\% & 9.77\% & 7.85\% & 0.38\% \\
\multicolumn{2}{c}{\textbf{Boost over MAE}} & 14.27\% & 3.43\% & 8.05\% & 1.94\% & 5.03\% & 2.36\% \\ \bottomrule
\end{tabular}
\end{table*}

\begin{table*}[]
\caption{Detailed results on six datasets with comprehensive evaluation metrics
}
\label{tab:FD}
\resizebox{\textwidth}{!}{%
\begin{tabular}{@{}lclllllll@{}}
\toprule
\textbf{Dataset} & \textbf{Pretrain} & \textbf{Method} & \multicolumn{1}{c}{\textbf{Accuracy}} & \multicolumn{1}{c}{\textbf{Precision}} & \multicolumn{1}{c}{\textbf{Recall}} & \multicolumn{1}{c}{\textbf{F1}} & \multicolumn{1}{c}{\textbf{AUROC}} & \multicolumn{1}{c}{\textbf{AUPRC}} \\ \midrule
\multirow{5}{*}{\textbf{FD}} & \multirow{2}{*}{\textbf{w/o}} & \textbf{SimCLR} & 0.6114 $\pm$ 0.0105 & 0.5224 $\pm$ 0.0081 & 0.6114 $\pm$ 0.0105 & 0.5460 $\pm$ 0.0141 & 0.7159 $\pm$ 0.0050 & 0.5299 $\pm$ 0.0068 \\
 &  & \textbf{MAE/CoGenT} & 0.9128 $\pm$ 0.0110 & 0.8711 $\pm$ 0.0124 & 0.9128 $\pm$ 0.0110 & 0.8894 $\pm$ 0.0117 & 0.9706 $\pm$ 0.0070 & 0.9536 $\pm$ 0.0091 \\ \cmidrule(l){2-9} 
 & \multirow{3}{*}{\textbf{w/}} & \textbf{SimCLR} & $0.6850  \pm 0.0328$ & 0.5958 $\pm$ 0.0367 & 0.6850 $\pm$ 0.0328 & 0.6063 $\pm$ 0.0182 & 0.7633 $\pm$ 0.0144 & 0.6073 $\pm$ 0.0299 \\
 &  & \textbf{MAE} & 0.8703 $\pm$ 0.0071 & 0.8275 $\pm$ 0.0039 & 0.8703 $\pm$ 0.0071 & 0.8447 $\pm$ 0.0045 & 0.9416 $\pm$ 0.0082 & 0.9179 $\pm$ 0.0128 \\
 &  & \textbf{CoGenT} & \textbf{0.9737 $\pm$ 0.0018} & \textbf{0.9576 $\pm$ 0.0054} & \textbf{0.9737 $\pm$ 0.0018} & \textbf{0.9652 $\pm$ 0.0039} & \textbf{0.9973 $\pm$ 0.0003} & \textbf{0.9968 $\pm$ 0.0004} \\ \midrule
\multirow{5}{*}{\textbf{FordA}} & \multirow{2}{*}{\textbf{w/o}} & \textbf{SimCLR} & 0.7249 $\pm$ 0.0053 & 0.7247 $\pm$ 0.0053 & 0.7249 $\pm$ 0.0053 & 0.7244 $\pm$ 0.0052 & 0.7937 $\pm$ 0.0076 & 0.7802 $\pm$ 0.0074 \\
 &  & \textbf{MAE/CoGenT} & 0.8993 $\pm$ 0.0059 & 0.9007 $\pm$ 0.0062 & 0.8993 $\pm$ 0.0059 & 0.8993 $\pm$ 0.0065 & 0.9617 $\pm$ 0.0021 & 0.9607 $\pm$ 0.0027 \\ \cmidrule(l){2-9} 
 & \multirow{3}{*}{\textbf{w/}} & \textbf{SimCLR} & 0.6742 $\pm$ 0.0105 & 0.6749 $\pm$ 0.0106 & 0.6742 $\pm$ 0.0105 & 0.6741 $\pm$ 0.0107 & 0.7340 $\pm$ 0.0058 & 0.7200 $\pm$ 0.0056 \\
 &  & \textbf{MAE} & 0.8828 $\pm$ 0.0033 & 0.8833 $\pm$ 0.0023 & 0.8828 $\pm$ 0.0033 & 0.8828 $\pm$ 0.0030 & 0.9481 $\pm$ 0.0045 & 0.9448 $\pm$ 0.0043 \\
 &  & \textbf{CoGenT} & \textbf{0.9129 $\pm$ 0.0034} & \textbf{0.9138 $\pm$ 0.0037} & \textbf{0.9129 $\pm$ 0.0034} & \textbf{0.9131 $\pm$ 0.0034} & \textbf{0.9715 $\pm$ 0.0015} & \textbf{0.9699 $\pm$ 0.0019} \\ \midrule
\multirow{5}{*}{\textbf{ClCon}} & \multirow{2}{*}{\textbf{w/o}} & \textbf{SimCLR} & 0.3761 $\pm$ 0.0346 & 0.3875 $\pm$ 0.0723 & 0.3761 $\pm$ 0.0346 & 0.3194 $\pm$ 0.1057 & 0.5504 $\pm$ 0.0294 & 0.3908 $\pm$ 0.0358 \\
 &  & \textbf{MAE/CoGenT} & 0.3944 $\pm$ 0.0108 & \textbf{0.4730} $\pm$ \textbf{0.0549} & 0.3944 $\pm$ 0.0108 & 0.3862 $\pm$ 0.0206 & 0.5766 $\pm$ 0.0061 & 0.4251 $\pm$ 0.0049 \\ \cmidrule(l){2-9} 
 & \multirow{3}{*}{\textbf{w/}} & \textbf{SimCLR} & 0.3740 $\pm$ 0.0312 & 0.3457 $\pm$ 0.1153 & 0.3740 $\pm$ 0.0312 & 0.3282 $\pm$ 0.1003 & 0.5481 $\pm$ 0.0241 & 0.4007 $\pm$ 0.0323 \\
 &  & \textbf{MAE} & 0.3954 $\pm$ 0.0058 & 0.4597 $\pm$ 0.0374 & 0.3954 $\pm$ 0.0058 & 0.3889 $\pm$ 0.0106 & 0.5806 $\pm$ 0.0073 & \textbf{0.4290 $\pm$ 0.0059} \\
 &  & \textbf{CoGenT} & \textbf{0.4223 $\pm$ 0.0076} & 0.4312 $\pm$ 0.0127 & \textbf{0.4223 $\pm$ 0.0076} & \textbf{0.4202 $\pm$ 0.0090} & \textbf{0.5944 $\pm$ 0.0048} & 0.4226 $\pm$ 0.0117 \\ \midrule
\multirow{5}{*}{\textbf{PTB}} & \multirow{2}{*}{\textbf{w/o}} & \textbf{SimCLR} & 0.5559 $\pm$ 0.0192 & 0.5353 $\pm$ 0.0133 & 0.5559 $\pm$ 0.0192 & 0.5384 $\pm$ 0.0162 & 0.7061 $\pm$ 0.0214 & 0.5572 $\pm$ 0.0036 \\
 &  & \textbf{MAE/CoGenT} & 0.6108 $\pm$ 0.0146 & \textbf{0.5915 $\pm$ 0.0225} & 0.6108 $\pm$ 0.0146 & 0.5971 $\pm$ 0.0179 & 0.7104 $\pm$ 0.0257 & \textbf{0.5963 $\pm$ 0.0177} \\ \cmidrule(l){2-9} 
 & \multirow{3}{*}{\textbf{w/}} & \textbf{SimCLR} & 0.5787 $\pm$ 0.0445 & 0.5453 $\pm$ 0.0191 & 0.5787 $\pm$ 0.0445 & 0.5495 $\pm$ 0.0234 & 0.6734 $\pm$ 0.0605 & 0.5532 $\pm$ 0.0130 \\
 &  & \textbf{MAE} & 0.6050 $\pm$ 0.0086 & 0.5837 $\pm$ 0.0072 & 0.6050 $\pm$ 0.0086 & 0.5917 $\pm$ 0.0067 & 0.6864 $\pm$ 0.0205 & 0.5898 $\pm$ 0.0068 \\
 &  & \textbf{CoGenT} & \textbf{0.6409 $\pm$ 0.0291} & 0.5891 $\pm$ 0.0131 & \textbf{0.6409 $\pm$ 0.0291} & \textbf{0.6032 $\pm$ 0.0164} & \textbf{0.7333 $\pm$ 0.0287} & 0.5874 $\pm$ 0.0164 \\ \midrule
\multirow{5}{*}{\textbf{ElecD}} & \multirow{2}{*}{\textbf{w/o}} & \textbf{SimCLR} & 0.5306 $\pm$ 0.0130 & 0.5367 $\pm$ 0.0075 & 0.5306 $\pm$ 0.0130 & 0.5301 $\pm$ 0.0091 & 0.8291 $\pm$ 0.0059 & 0.5277 $\pm$ 0.0102 \\
 &  & \textbf{MAE/CoGenT} & 0.5717 $\pm$ 0.0123 & 0.5796 $\pm$ 0.0147 & 0.5717 $\pm$ 0.0123 & 0.5647 $\pm$ 0.0130 & \textbf{0.8711 $\pm$ 0.0083} & 0.5977 $\pm$ 0.0141 \\ \cmidrule(l){2-9} 
 & \multirow{3}{*}{\textbf{w/}} & \textbf{SimCLR} & 0.5506 $\pm$ 0.0078 & 0.5605 $\pm$ 0.0067 & 0.5506 $\pm$ 0.0078 & 0.5515 $\pm$ 0.0065 & 0.8360 $\pm$ 0.0069 & 0.5395 $\pm$ 0.0088 \\
 &  & \textbf{MAE} & 0.5700 $\pm$ 0.0094 & 0.5790 $\pm$ 0.0099 & 0.5700 $\pm$ 0.0094 & 0.5663 $\pm$ 0.0085 & 0.8686 $\pm$ 0.0074 & 0.5921 $\pm$ 0.0154 \\
 &  & \textbf{CoGenT} & \textbf{0.5998 $\pm$ 0.0043} & \textbf{0.6140 $\pm$ 0.0064} & \textbf{0.5998 $\pm$ 0.0043} & \textbf{0.5948 $\pm$ 0.0031} & 0.8699 $\pm$ 0.0056 & \textbf{0.6249 $\pm$ 0.0116} \\ \midrule
\multirow{5}{*}{\textbf{HAR}} & \multirow{2}{*}{\textbf{w/o}} & \textbf{SimCLR} & 0.8773 $\pm$ 0.0027 & 0.8788 $\pm$ 0.0030 & 0.8773 $\pm$ 0.0027 & 0.8757 $\pm$ 0.0022 & 0.9839 $\pm$ 0.0017 & 0.9339 $\pm$ 0.0071 \\
 &  & \textbf{MAE/CoGenT} & 0.8914 $\pm$ 0.0180 & 0.8940 $\pm$ 0.0192 & 0.8927 $\pm$ 0.0174 & 0.8889 $\pm$ 0.0202 & 0.9807 $\pm$ 0.0104 & 0.9423 $\pm$ 0.0208 \\ \cmidrule(l){2-9} 
 & \multirow{3}{*}{\textbf{w/}} & \textbf{SimCLR} & 0.8943 $\pm$ 0.0062 & 0.8962 $\pm$ 0.0072 & 0.8943 $\pm$ 0.0062 & 0.8931 $\pm$ 0.0062 & \textbf{0.9866 $\pm$ 0.0013} & 0.9419 $\pm$ 0.0059 \\
 &  & \textbf{MAE} & 0.8775 $\pm$ 0.0093 & 0.8772 $\pm$ 0.0083 & 0.8775 $\pm$ 0.0093 & 0.8758 $\pm$ 0.0094 & 0.9769 $\pm$ 0.0035 & 0.9236 $\pm$ 0.0070 \\
 &  & \textbf{CoGenT} & \textbf{0.8976 $\pm$ 0.0076} & \textbf{0.8968 $\pm$ 0.0066} & \textbf{0.8976 $\pm$ 0.0076} & \textbf{0.8965 $\pm$ 0.0074} & 0.9861 $\pm$ 0.0004 & \textbf{0.9438 $\pm$ 0.0030} \\ \bottomrule
\end{tabular}%
}
\end{table*}

% \subsubsection{Overall results}
\header{Overall results}
We present the comparison with baselines on six datasets in Table~\ref{tab:results}. 
From the results, we can clearly observe that \name\ achieves the best F1-score and significantly outperforms both SimCLR and MAE across all datasets. Notably, on FD, \name\ achieves 0.9652, outperforming SimCLR (0.6063) by \textbf{59.2\%} in relative F1 score and MAE (0.8447) by \textbf{14.27\%}. 
On FordA, \name\ reaches 0.9131, showing a \textbf{35.45\%} improvement over SimCLR and a \textbf{3.43\%} gain over MAE. 
Similar trends are observed on other datasets, including ClCon (0.4202), PTB (0.6032), ElecD (0.5948), and HAR (0.8965), confirming the generalizability of \name\ across a wide range of time series tasks.  
Overall, \name\ outperforms the solely contrastive SSL and generative models by up to 59.2\% and 14.27\% in F1 score, respectively.

\subsubsection{Detailed results}
Next, we report the detailed results in Table~\ref{tab:FD}, including accuracy, precision, recall, F1 score, AUROC (area under the ROC curve), and AUPRC (area under the Precision-Recall curve).
% We take the FD as an example (Table~\ref{tab:FD}) to provide comprehensive evaluation metrics,  
Taking FD as an example, beyond \name's superior performance, we observe that pretraining can degrade MAE's performance compared to training from scratch. One potential reason is that generative SSL needs large-scale datasets to learn meaningful data distributions. When pretrained on smaller datasets, MAE may converge to suboptimal initializations. While verification on larger datasets would test this hypothesis, the quadratic computational cost leads us to defer this analysis to future work.

% Apart from the superior performance of \name, we also found that pre-training may cause the degrading of MAE. Although the key focus of this work is \name, we try to explain the observation of MAE's performance: Generative SSL has a high requirment on data scales, so that the pretraining stage of MAE provides a worse parameter initialization than directly train from scratch. One way to verify it is to test the model on a much larger dataset. however, this will quadratically increase the computational cost, we gonna leave it to the future work.  

% \subsubsection{Results Analysis}
\header{Results Analysis}
The substantial performance gap between our method and the baselines can be attributed to fundamental differences in how contrastive and generative approaches handle time series data. 

For contrastive learning methods like SimCLR, we observe significant performance degradation on datasets with high intra-class similarity, particularly in FD and FordA, which contain mechanical vibration signals with extremely high frequency components. 
The core challenge is how to define appropriate negative samples: when samples from the same class exhibit minimal variation, the contrastive objective may inadvertently push similar instances apart, thereby harming the learned representations. 
This effect is especially pronounced in FD's case, where the 16 KHz sampling rate combined with short 1280-sample segments (just 0.08 seconds of data) results in remarkably similar signal patterns across samples. While increasing the segment length could potentially improve discrimination, this approach is computationally prohibitive due to the quadratic complexity of self-attention and known limitations of Transformers in processing very long sequences.

The generative approach of MAE demonstrates superior performance on these same datasets by focusing on reconstructing the overall signal distribution rather than discriminating between individual instances. The inherent similarity between patches actually benefits MAE's reconstruction objective, to the extent that in some cases the non-pretrained version outperforms its pretrained counterpart. However, MAE's effectiveness remains constrained by dataset scale, as evidenced by its smallest performance gap to our method (1.94\%) happening on PTB, our largest dataset. This suggests that while MAE can effectively model local signal distributions, it requires substantial data to learn robust global representations.

The comparative analysis of non-pretrained versions reveals additional architectural insights. MAE's consistent outperformance of SimCLR, despite both using fundamentally similar Transformer backbones, highlights the importance of a proper framework for time series data. The key differentiators, including multivariate patching and patch-level masking, could be particularly effective architectural choices for time series representation learning. 
% We will discuss more in the Discussion.

\begin{figure*}[h]
    \centering
    \includegraphics[width=\linewidth]{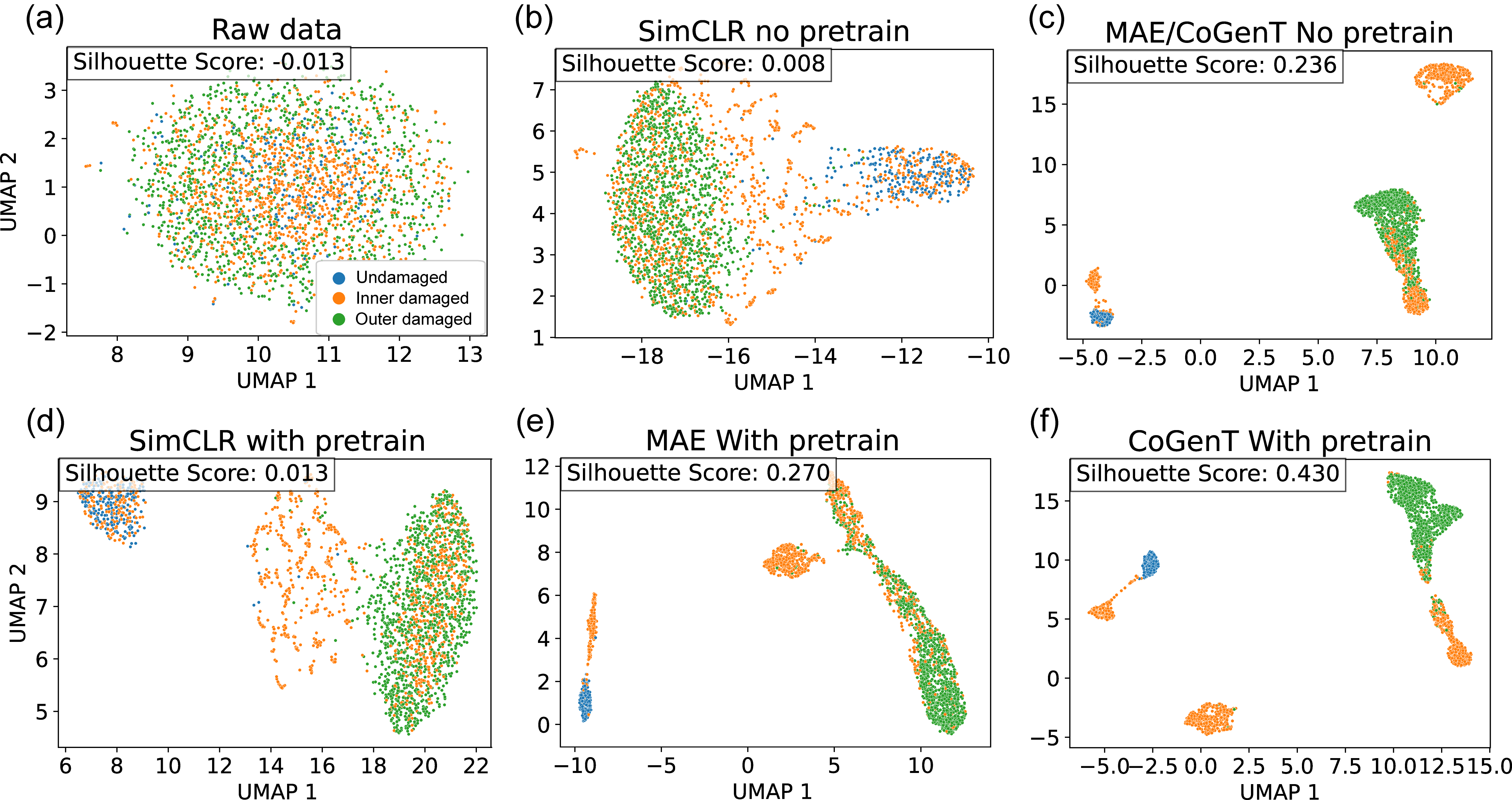}
    \caption{Representation visualization (FD; 3-classes) comparing the embeddings from raw signals, non-pretrained SimCLR and MAE, along with pretrained SimCLR, MAE, and \name. We also provide the Silhouette score as a quantitative measurement.}
    \label{fig:visualization}
\end{figure*}
 
% \subsubsection{Summary}
\header{Summary}
The superior performance of \name\ stems from its unified framework that fundamentally integrates contrastive and generative self-supervised learning. This creates a synergistic effect where each component addresses fundamental limitations of the other: 1) the contrastive objective ensures discriminative power by learning to separate dissimilar samples while pulling together similar ones, which is essential for downstream classification tasks; 2) simultaneously, the generative objective imposes structural regularization on the latent space, enforcing robustness by requiring the model to capture the complete data distribution rather than just discriminative features.

By jointly optimizing these complementary objectives, \name\ achieves three key advantages: 
1) it overcomes contrastive learning's sensitivity to high intra-class similarity by leveraging the generative component's distributional awareness; 2) it mitigates generative methods' dependence on large datasets through the contrastive component's sample-efficient discrimination; 3) it demonstrates consistent robustness to noise and limited data scenarios by combining both approaches' strengths.

% 1) mitigates contrastive learning's sensitivity to high intra-class similarity, 2) reduces generative methods' data dependence, and 3) enhances robustness to noise and small datasets—explaining \name's consistent outperformance.  

% \subsection{Loss Balancer Comparasion}
% \xiang{Simplely compare 1) fixed coefficients, no balancer; 2) Gradbalancer; 3) your balancer on one dataset. 4) loss curves of rec loss and contrast loss, in above three situations}

\subsection{Representation Visualization}
We also visualize the learned feature representations to validate the effectiveness of \name. 
Using the FD dataset as an example, we compare six representations: the raw signal, features learned by non-pretrained SimCLR and MAE, and pretrained representations from SimCLR, MAE, and \name. 
For the raw data, we use the signal directly, while for the five models, we extract features from the second-to-last layer of the classifier.
Since the representations are high-dimensional, we employ UMAP to project them into a 2-D space, visualized in Figure~\ref{fig:visualization}. We further quantify cluster separation using the Silhouette Score (SS).

The raw signal exhibits no discernible clustering (SS = -0.013), appearing as an overlapping dot cloud, which aligns with expectations given its noisy nature.
Although the non-pretrained SimCLR (SS = 0.008) shows negligible improvement, it successfully separates undamaged samples (right cluster) from damaged bearings while failing to distinguish between inner and outer damage. Pretrained SimCLR (SS = 0.013) performs marginally better.

In contrast, MAE and \name\ demonstrate significant improvements with pretraining. The non-pretrained MAE obtains an SS of 0.236, which rises to 0.27 after pretraining. While MAE partially separates inner (yellow) and outer (green) damage clusters, the boundaries remain imprecise. 
Notably, the proposed \name\ attains the highest SS (0.43), outperforming SimCLR and MAE by a substantial margin. Its visualization reveals a clear separation of inner and outer damage samples (top-right block), highlighting its superior representation learning capability.
 
\begin{table*}[h]
\centering
\caption{Ablation Study (FD). The $\mathcal{L}_{r_i}$ and $\mathcal{L}_{r'_i}$ denote the reconstruction loss of the original sample  $\bm{x}_i$
and the augmented sample $\bm{x}'_i$, respectively.
In short, $\mathcal{L}_{r} = (\mathcal{L}_{r_i} + \mathcal{L}_{r'_i})/2$. 
}
\label{tab:ablation_study}
\begin{tabular}{@{}lllllll@{}}
\toprule
\textbf{Loss} & \multicolumn{1}{c}{\textbf{Accuracy}} & \multicolumn{1}{c}{\textbf{Precision}} & \multicolumn{1}{c}{\textbf{Recall}} & \multicolumn{1}{c}{\textbf{F1}} & \multicolumn{1}{c}{\textbf{AUROC}} & \multicolumn{1}{c}{\textbf{AUPRC}} \\ \midrule
$L_{r_i}$ & 0.8751 $\pm$ 0.0058 & 0.8333 $\pm$ 0.0043 & 0.8751 $\pm$ 0.0058 & 0.8517 $\pm$ 0.0038 & 0.9403 $\pm$ 0.0078 & 0.9135 $\pm$ 0.0118 \\
$L_{r_i}$ + $L_{r'_i}$ & 0.8841 $\pm$ 0.0111 & 0.8532 $\pm$ 0.0155 & 0.8841 $\pm$ 0.0111 & 0.8644 $\pm$ 0.0157 & 0.9555 $\pm$ 0.0048 & 0.9370 $\pm$ 0.0026 \\
$L_{r_i}$ + $L_{r'_i}$ + $L_c$ & 0.9737 $\pm$ 0.0018 & 0.9576 $\pm$ 0.0054 & 0.9737 $\pm$ 0.0018 & 0.9652 $\pm$ 0.0039 & 0.9973 $\pm$ 0.0003 & 0.9967 $\pm$ 0.0004 \\ \bottomrule
\end{tabular}
\end{table*}

\subsection{Ablation Study}
The proposed \name\ is a unified framework comprising two key components: a contrastive module and a generative module. To validate the effectiveness of each component, we conduct an ablation study. As shown in Eq~\ref{eq:total_loss}, the two modules are represented by $\mathcal{L}_{c}$ and $\mathcal{L}_{r}$. In which, the reconstruction loss contains two parts: reconstructing the original sample ($\bm{x}_i, \hat{\bm{x}_i}$) and reconstructing the augmented sample ($\bm{x}'_i, \hat{\bm{x}'_i}$), let's denote them by $\mathcal{L}_{r_i}$ and $\mathcal{L}_{r'_i}$, respectively. To clarify, we have $\mathcal{L}_{r} = (\mathcal{L}_{r_i} + \mathcal{L}_{r'_i})/2$.

We report the ablation study results in Table~\ref{tab:ablation_study}. The model achieves an F1 score of 0.8517 when trained solely with the reconstruction loss of the original sample $\bm{x}_i$. 
Incorporating the reconstruction loss of augmented views yields a marginal improvement (F1 = 0.8644). However, when the contrastive loss is added, the F1 score rises significantly to 0.9652, demonstrating that the contrastive module plays a crucial role in enhancing model performance. 

\section{Discussions}
While \name\ demonstrates significant performance improvements as the first framework to unify contrastive and generative SSL for multivariate time series, several limitations and future opportunities need discussion. 

The choice of SimCLR and MAE as representative baselines, while well-justified for this initial investigation, presents opportunities for extension. Recent advances in both contrastive and generative SSL, as surveyed in Section~\ref{sec:related_work}, suggest promising directions for enhancement. Particularly compelling would be the integration of more sophisticated MAE variants employing hierarchical transformer architectures, which could better capture multi-scale temporal patterns. Similarly, the contrastive component could be augmented with additional discrimination tasks, such as incorporating subject-level differentiation alongside the current sample-level contrastive objective, potentially yielding richer representations.

In terms of model architecture, our deep fusion approach in embedding space ($h$) represents one of several possible integration strategies. Alternative designs employing parallel processing branches with late-stage concatenation, or multi-phase fusion approaches, might offer different trade-offs between performance and modularity. While we aware that such alternatives may not match the representational synergy of our proposed deep fusion method, systematic empirical comparison would provide valuable insights for the field and could reveal context-dependent advantages.

The computational requirements of \name\ naturally exceed those of its individual components, presenting both challenges and research opportunities. Processing both original and augmented views necessarily doubles the input data volume, while the joint optimization of contrastive and generative losses increases the computational cost by approximately 1.8-2 times. This overhead, while necessary, appears inherent to the hybrid SSL paradigm and motivates investigation into optimization techniques specifically designed for such multi-objective frameworks.

Finally, the relationship between dataset scale and model performance requires deeper investigation. While PTB provides a substantial testbed, the field would benefit from evaluation on even larger-scale time series collections ($>$1M samples) to properly characterize scaling laws. 
One particular interest would be identifying the critical dataset size at which generative pretraining becomes consistently beneficial across domains, as well as quantifying how this threshold varies with the proposed hybrid approach. Such analysis would not only validate our current findings but also provide practical guidance for real-world applications.

% First, this model only considered SimCLR and MAE as representatives of contrastive and generative SSL. We are aware of dozens of new variations of contrative and generative SSL as mentioned in Section~\ref{sec:related_work}. One promising future work is to upgrade the encoder into more advanced MAE encoders, and increase the calculation of contrastive loss functions (e.g., including subject discrimination upon the current sample discrimination loss). 

% Moreover, now we deeply integrate the contrastive (in embedding space $h$) and the reconstruction. There might be other integration methods, such as two input branches for contrastrie and generative seperately, then concatenate their embeddings together. Such in-parallal integration could also work, but likely underperform the proposed \name\ with deep feature fusion. However, this still worth to explore. 

% Thrid, comparing with a standard MAE, \name\ receives both original sample and augmented view, in other words, the datasize has been doubled so that the computing complexity linearly increased. Accoridng to our knowledge, since we are fusing two different paradigms, it will inevitablly increase the computing cost. We will leave this as an opening challenge. 

% Forth, we assume the MAE gonna work better when working on a larger dataset. Although PTB is already a large time series dataset, we will retrival other gaint scale datasets to validate the scaling law on MAE and \name. 

\section{Conclusion}
We presented \name, a unified framework that effectively combines contrastive and generative SSL for time series analysis. This unified design explains \name's significant and consistent performance gains across all evaluated datasets, particularly in challenging conditions where standalone methods struggle. 
This work opens new directions for efficient hybrid architectures and broader applications beyond classification. We release all code to support future research in unified SSL paradigms for temporal data.

 % The framework's flexibility suggests promising potential for extension to other temporal data learning tasks.

% \section*{Acknowledgment}

\bibliographystyle{ieeetr}
\bibliography{ref.bib}

\begin{thebibliography}{10}

\bibitem{liu2021self}
X.~Liu, F.~Zhang, Z.~Hou, L.~Mian, Z.~Wang, J.~Zhang, and J.~Tang, ``Self-supervised learning: Generative or contrastive,'' {\em IEEE Transactions on Knowledge and Data Engineering}, vol.~35, no.~1, pp.~857--876, 2021.

\bibitem{zhang2024self}
K.~Zhang, Q.~Wen, C.~Zhang, R.~Cai, M.~Jin, Y.~Liu, J.~Y. Zhang, Y.~Liang, G.~Pang, D.~Song, {\em et~al.}, ``Self-supervised learning for time series analysis: Taxonomy, progress, and prospects,'' {\em IEEE transactions on pattern analysis and machine intelligence}, 2024.

\bibitem{chen2020simple}
T.~Chen, S.~Kornblith, M.~Norouzi, and G.~Hinton, ``A simple framework for contrastive learning of visual representations,'' in {\em International conference on machine learning}, pp.~1597--1607, PMLR, 2020.

\bibitem{yang2022timeclr}
X.~Yang, Z.~Zhang, and R.~Cui, ``Timeclr: A self-supervised contrastive learning framework for univariate time series representation,'' {\em Knowledge-Based Systems}, vol.~245, p.~108606, 2022.

\bibitem{yue2022ts2vec}
Z.~Yue, Y.~Wang, J.~Duan, T.~Yang, C.~Huang, Y.~Tong, and B.~Xu, ``Ts2vec: Towards universal representation of time series,'' in {\em Proceedings of the AAAI Conference on Artificial Intelligence}, vol.~36, pp.~8980--8987, 2022.

\bibitem{zhangself}
X.~Zhang, Z.~Zhao, T.~Tsiligkaridis, and M.~Zitnik, ``Self-supervised contrastive pre-training for time series via time-frequency consistency,'' in {\em Advances in Neural Information Processing Systems}, 2022.

\bibitem{he2022masked}
K.~He, X.~Chen, S.~Xie, Y.~Li, P.~Doll{\'a}r, and R.~Girshick, ``Masked autoencoders are scalable vision learners,'' in {\em Proceedings of the IEEE/CVF Conference on Computer Vision and Pattern Recognition}, pp.~16000--16009, 2022.

\bibitem{feichtenhofer2022masked}
C.~Feichtenhofer, Y.~Li, K.~He, {\em et~al.}, ``Masked autoencoders as spatiotemporal learners,'' {\em Advances in neural information processing systems}, vol.~35, pp.~35946--35958, 2022.

\bibitem{zerveas2021tsmae}
G.~Zerveas, S.~Jayaraman, D.~Patel, A.~Bhamidipaty, and C.~Eickhoff, ``{TS-MAE}: A masked autoencoder for time series representation learning,'' {\em arXiv preprint arXiv:2106.11046}, 2021.

\bibitem{tang2022mtsmae}
P.~Tang and X.~Zhang, ``{MTSMAE}: Masked autoencoders for multivariate time-series forecasting,'' {\em arXiv preprint arXiv:2210.02199}, 2022.

\bibitem{cheng2023timemae}
Z.~Cheng, M.~Ren, Y.~Qin, and Q.~Liu, ``{TimeMAE}: Self-supervised representations of time series with decoupled masked autoencoders,'' {\em arXiv preprint arXiv:2303.00320}, 2023.

\bibitem{henaff2020data}
O.~Henaff, ``Data-efficient image recognition with contrastive predictive coding,'' in {\em International conference on machine learning}, pp.~4182--4192, PMLR, 2020.

\bibitem{eldele2021time}
E.~Eldele, M.~Ragab, Z.~Chen, M.~Wu, C.-K. Kwoh, X.~Li, and C.~Guan, ``Time-series representation learning via temporal and contextual contrasting,'' {\em arXiv preprint arXiv:2106.14112}, 2021.

\bibitem{woocost2022}
G.~Woo, C.~Liu, D.~Sahoo, A.~Kumar, and S.~Hoi, ``Cost: Contrastive learning of disentangled seasonal-trend representations for time series forecasting,'' in {\em International Conference on Learning Representations}, 2022.

\bibitem{leesoft2022}
S.~Lee, T.~Park, and K.~Lee, ``Soft contrastive learning for time series,'' in {\em The Twelfth International Conference on Learning Representations}, 2022.

\bibitem{shamba2024dynamic}
A.-K. Shamba, K.~Bach, and G.~Taylor, ``Dynamic contrastive learning for time series representation,'' {\em arXiv preprint arXiv:2410.15416}, 2024.

\bibitem{li2023timae}
Z.~Li, Z.~Rao, L.~Pan, P.~Wang, and Z.~Xu, ``{Ti-MAE}: Self-supervised masked time series autoencoders,'' {\em arXiv preprint arXiv:2301.08871}, 2023.

\bibitem{liang2022self}
D.~Liang, J.~Wang, X.~Gao, J.~Wang, X.~Zhao, and L.~Wang, ``Self-supervised pretraining isolated forest for outlier detection,'' in {\em 2022 International Conference on Big Data, Information and Computer Network (BDICN)}, pp.~306--310, IEEE, 2022.

\bibitem{fang2024tfmae}
X.~Fang, K.~Xu, K.~Zheng, J.~Zhou, and X.~Lin, ``Temporal-frequency masked autoencoders for time series anomaly detection,'' in {\em Proceedings of the IEEE International Conference on Data Engineering ({ICDE})}, 2024.

\bibitem{gao2023hscmae}
W.~Gao, Y.~Zhang, Y.~Li, Y.~Liu, Q.~Li, and W.~Wang, ``{HSC-MAE}: Self-supervised learning-based time series classification via hierarchical sparse convolutional masked-autoencoder,'' {\em arXiv preprint arXiv:2305.12345}, 2023.

\bibitem{zhang2022maeeeg}
H.~Zhang, Z.~Zhang, L.~Wang, Z.~Li, J.~Zhang, and H.~He, ``{MAEEG}: Masked auto-encoder for {EEG} representation learning,'' {\em arXiv preprint arXiv:2207.01500}, 2022.

\bibitem{jiang2023stmae}
Y.~Jiang, S.~Zhang, and Z.~Li, ``Revealing the power of masked autoencoders in traffic forecasting,'' {\em arXiv preprint arXiv:2306.00100}, 2023.

\bibitem{dong2023simmtm}
J.~Dong, H.~Wu, H.~Zhang, L.~Zhang, J.~Wang, and M.~Long, ``{SimMTM}: A simple pre-training framework for masked time-series modeling,'' {\em arXiv preprint arXiv:2302.00861}, 2023.

\bibitem{zha2022extramae}
D.~Zha, X.~H. Yu, Y.~Zhang, J.~Hu, and X.~Hu, ``Extramae: Time series generation with masked autoencoder,'' {\em arXiv preprint arXiv:2201.07006}, 2022.

\bibitem{huang2023contrastive}
Z.~Huang, X.~Jin, C.~Lu, Q.~Hou, M.-M. Cheng, D.~Fu, X.~Shen, and J.~Feng, ``Contrastive masked autoencoders are stronger vision learners,'' {\em IEEE Transactions on Pattern Analysis and Machine Intelligence}, vol.~46, no.~4, pp.~2506--2517, 2023.

\bibitem{li2021generative}
B.~Li, Y.~Li, and K.~W. Eliceiri, ``Generative and contrastive based self-supervised learning model for histopathology image analysis,'' {\em Medical Image Analysis}, vol.~73, p.~102144, 2021.

\bibitem{zhang2021contrastive}
P.~Zhang, F.~Wang, and Y.~Zheng, ``Contrastive generative self-supervised learning for 3d medical image segmentation,'' {\em Medical Image Analysis}, vol.~73, p.~102142, 2021.

\bibitem{chen2022can}
X.~Chen, S.~Xie, and K.~He, ``Can: A simple, efficient and scalable contrastive masked autoencoder framework for learning visual representations,'' {\em arXiv preprint arXiv:2205.09843}, 2022.

\bibitem{wan2020contrastive}
S.~Wan, S.~Pan, J.~Yang, and C.~Gong, ``Contrastive and generative graph convolutional networks for graph-based semi-supervised learning,'' {\em arXiv preprint arXiv:2009.07111}, 2020.

\bibitem{wang2022hicmae}
Y.~Wang, H.~Wu, J.~Li, and J.~Wang, ``Hicmae: Hierarchical contrastive masked autoencoder for self-supervised audio-visual emotion recognition,'' {\em IEEE Transactions on Multimedia}, 2022.

\bibitem{liu2023self}
Z.~Liu, A.~Alavi, M.~Li, and X.~Zhang, ``Self-supervised contrastive learning for medical time series: A systematic review,'' {\em Sensors}, vol.~23, no.~9, p.~4221, 2023.

\bibitem{liu2024guidelines}
Z.~Liu, A.~Alavi, M.~Li, and X.~Zhang, ``Guidelines for augmentation selection in contrastive learning for time series classification,'' {\em arXiv preprint arXiv:2407.09336}, 2024.

\bibitem{nie2022time}
Y.~Nie, N.~H. Nguyen, P.~Sinthong, and J.~Kalagnanam, ``A time series is worth 64 words: Long-term forecasting with transformers,'' {\em arXiv preprint arXiv:2211.14730}, 2022.

\bibitem{wang2024medformer}
Y.~Wang, N.~Huang, T.~Li, Y.~Yan, and X.~Zhang, ``Medformer: A multi-granularity patching transformer for medical time-series classification,'' in {\em The Thirty-eighth Annual Conference on Neural Information Processing Systems}, 2024.

\bibitem{lessmeier2016condition}
C.~Lessmeier, J.~K. Kimotho, D.~Zimmer, and W.~Sextro, ``Condition monitoring of bearing damage in electromechanical drive systems by using motor current signals of electric motors: A benchmark data set for data-driven classification,'' in {\em PHM Society European Conference}, vol.~3, 2016.

\bibitem{zhang2022self}
X.~Zhang, Z.~Zhao, T.~Tsiligkaridis, and M.~Zitnik, ``Self-supervised contrastive pre-training for time series via time-frequency consistency,'' {\em NeurIPS}, vol.~35, pp.~3988--4003, 2022.

\bibitem{ucrarchive}
H.~A. Dau, A.~Bagnall, K.~Kamgar, C.-C.~M. Yeh, Y.~Zhu, S.~Gharghabi, C.~A. Ratanamahatana, and E.~Keogh, ``The ucr time series archive,'' 2018.

\bibitem{li2011fast}
L.~Li, {\em Fast algorithms for mining co-evolving time series}.
\newblock Carnegie Mellon University, 2011.

\bibitem{bousseljot1995nutzung}
R.~Bousseljot, D.~Kreiseler, and A.~Schnabel, ``Nutzung der ekg-signaldatenbank cardiodat der ptb {\"u}ber das internet,'' 1995.

\bibitem{goldberger2000physiobank}
A.~L. e.~a. Goldberger, ``Physiobank, physiotoolkit, and physionet: components of a new research resource for complex physiologic signals,'' {\em circulation}, vol.~101, no.~23, pp.~e215--e220, 2000.

\bibitem{wang2024evaluate}
Y.~Wang, T.~Li, Y.~Yan, W.~Song, and X.~Zhang, ``How to evaluate your medical time series classification?,'' {\em arXiv preprint arXiv:2410.03057}, 2024.

\bibitem{lines2011classification}
J.~Lines, A.~Bagnall, P.~Caiger-Smith, and S.~Anderson, ``Classification of household devices by electricity usage profiles,'' in {\em Intelligent Data Engineering and Automated Learning-IDEAL 2011: 12th International Conference, Norwich, UK, September 7-9, 2011. Proceedings 12}, pp.~403--412, Springer, 2011.

\bibitem{misc_human_activity_recognition_using_smartphones_240}
J.~Reyes-Ortiz, D.~Anguita, A.~Ghio, L.~Oneto, and X.~Parra, ``{Human Activity Recognition Using Smartphones}.'' UCI Machine Learning Repository, 2012.
\newblock {DOI}: https://doi.org/10.24432/C54S4K.

\end{thebibliography}
% \bibliography{ref.bib}

\end{document}